\documentclass[journal]{IEEEtran}
\usepackage{times}
\usepackage{latexsym}
\usepackage{url}
\usepackage{algorithm}
\usepackage{algcompatible}
\usepackage[noend]{algpseudocode}
\usepackage{amssymb}
\usepackage{multirow}
\usepackage{verbatim}
\usepackage{scrextend}
\usepackage{float}
\usepackage{subfig}
\usepackage{array}

\usepackage{mathtools}

\usepackage{tablefootnote}

\renewcommand{\arraystretch}{1.2}
\newcolumntype{C}[1]{>{\centering\let\newline\\\arraybackslash\hspace{0pt}}m{#1}}
\usepackage{tikz}
\usetikzlibrary{shapes,arrows}
\usepackage{amsmath,bm,times}
\usepackage{verbatim}
\usepackage{subfig}
\let\labelindent\relaxs
\usepackage{enumitem}

\usepackage{float}
\usetikzlibrary{positioning}
\usepackage{tkz-graph}

\usetikzlibrary{decorations.pathreplacing,intersections}
\usepackage{wrapfig}
\usepackage{pbox}
\usepackage{colortbl}
\usepackage{xcolor,soul,framed}
\colorlet{shadecolor}{yellow}
\usepackage{eqparbox}
\usepackage{url}
\usepackage{wrapfig}
\usepackage{pifont}
\usepackage{amssymb}

\usepackage{booktabs}
\usepackage{arydshln}
\usepackage{todonotes}

\newcommand{\model}{\texttt{EmpRes}}

\begin{document}

\title{Sentiment-guided Commonsense-aware Response Generation for Mental Health Counseling}

\author{Aseem Srivastava\textsuperscript{$1$}, Gauri Naik\textsuperscript{$1$}, Alison Cerezo\textsuperscript{$2,3$}, Tanmoy Chakraborty\textsuperscript{$4$}, Md. Shad Akhtar\textsuperscript{$1$}\\
{\em \textsuperscript{$1$}IIIT Delhi, India; \textsuperscript{$2$}University of California, Santa Barbara; \textsuperscript{$3$}Mpathic.ai; \textsuperscript{$4$}IIT Delhi, India}\\
{\em \{aseems, gaurin, shad.akhtar\}@iiitd.ac.in; acerezo@ucsb.edu; tanchak@iitd.ac.in}
}

\maketitle

\begin{abstract}
The crisis of mental health issues is escalating alarmingly. Effective counseling serves as a critical lifeline for individuals suffering from conditions like PTSD, stress, depression, etc. Therapists forge a crucial therapeutic bond with clients, steering them towards positivity. Unfortunately, the massive shortage of professionals, high costs, and mental health stigma pose significant barriers to consulting therapists. As a substitute, Virtual Mental Health Assistants (VMHAs) have emerged in the digital healthcare space. However, most existing VMHAs lack the commonsense to understand the nuanced sentiments of clients to generate effective responses. To this end, we propose \model, a novel sentiment-guided mechanism incorporating commonsense awareness for generating responses. By leveraging foundation models and harnessing commonsense knowledge, \model\ aims to generate responses that effectively shape the client's sentiment towards positivity. We evaluate the performance of \model\ on HOPE, a benchmark counseling dataset, and observe a remarkable performance improvement compared to the existing baselines across a suite of qualitative and quantitative metrics. Moreover, our extensive empirical analysis and human evaluation show that the generation ability of \model\ is well-suited and, in some cases, surpasses the gold standard. Further, we deploy \model\ as a chat interface for users seeking mental health support. We address the deployed system's effectiveness through an exhaustive user study with a significant positive response. Our findings show that 91\% of users find the system effective, 80\% express satisfaction, and over 85.45\% convey a willingness to continue using the interface and recommend it to others, demonstrating the practical applicability of \model\ in addressing the pressing challenges of mental health support, emphasizing user feedback, and ethical considerations in a real-world context.
\end{abstract}


\IEEEpeerreviewmaketitle

\section{Introduction}
Mental health issues are on the incline, demanding immediate attention. However, the accessibility to mental health professionals remains limited, exacerbating the challenges faced by individuals seeking support. In fact, according to a recent report by Mental Health America 2023, out of the staggering 50 million people affected by mental health issues, over 55\% did not receive the necessary treatment, highlighting the magnitude of the problem\footnote{https://mhanational.org/issues/state-mental-health-america}. To address the gap of experts and meet the requirements of support-seekers, virtual mental health assistants (VMHAs) come to the rescue. 
\begin{figure}[t]
  \centering
  \includegraphics[width=\columnwidth]{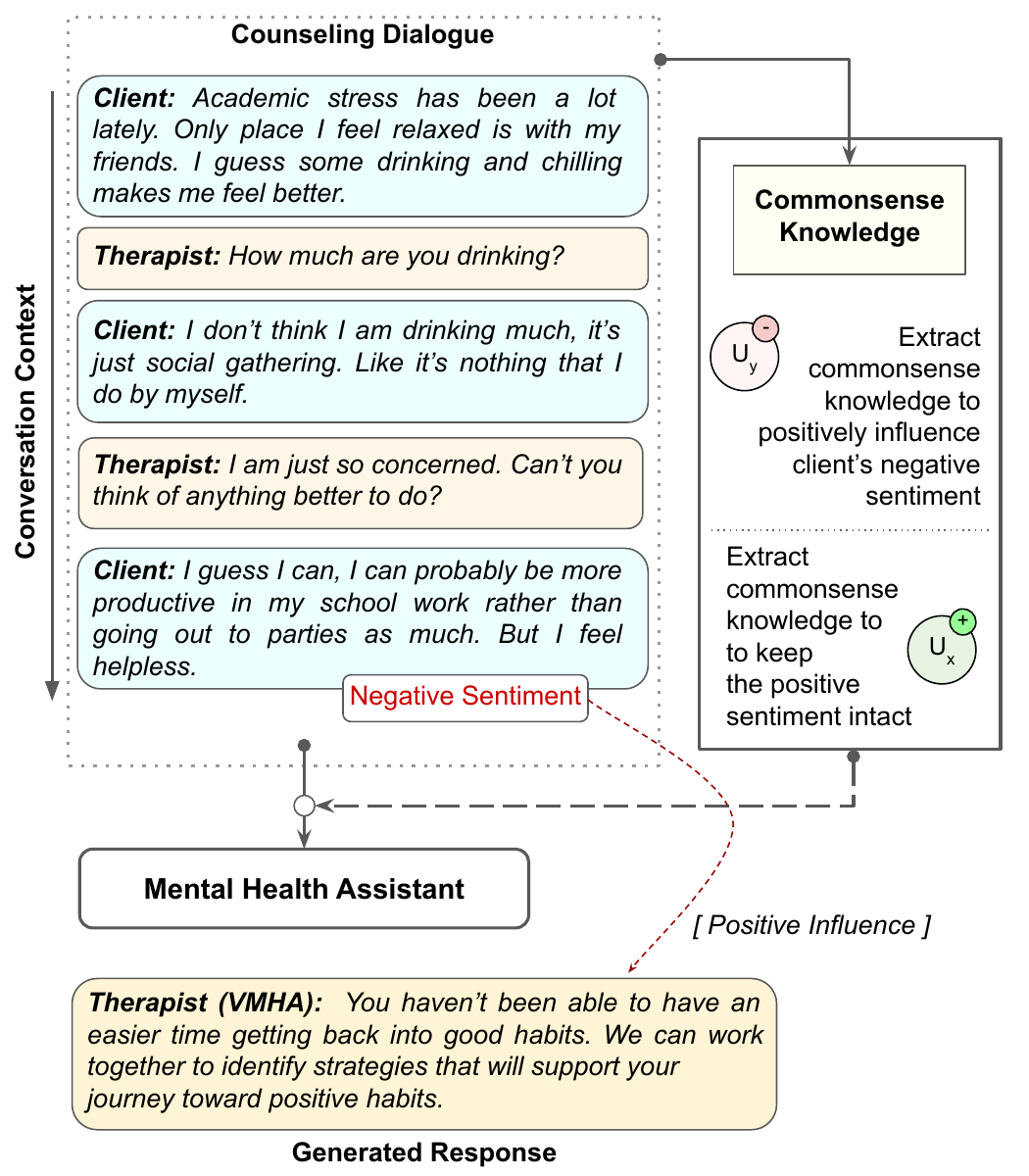}
  \caption{A sample counseling conversation in which the client expresses negative sentiment at last; the job of a VMHA is to generate a response to positively influence the client and contains rich commonsense knowledge.}
  \label{fig:page1example}
\end{figure}
The VMHA's ultimate goal is to provide effective responses that positively influence clients' sentiments to lead toward healthy counseling practices and ensure client stability throughout sessions.
Studies suggest prioritizing control over clients' sentimental feelings as the key focus for counselors working with new clients displaying negativity \cite{deangelis}. In a real-world setting, mental health experts proficiently employ their commonsense knowledge, along with active listening and validation skills, to understand clients' experiences and subsequently provide effective responses during counseling. To mimic such a counseling approach, our research aims to explore the benefit of commonsense knowledge as a means to aid VMHAs in understanding clients. To explain this better, a sample of the counseling session is shown in Figure \ref{fig:page1example}. As we observe, each counseling utterance is tagged with a sentiment label, which is utilized by the knowledge extractor in a way such that the mental health assistant generates a response that influences the client positively. Several VMHAs, including notable ones like Wysa\footnote{https://wysa.com} and Woebot\footnote{https://woebothealth.com}, are designed to function as mental health professionals. However, they currently lack the capability to primarily address clients' sentimental states and respond using commonsense knowledge within their conversational methodologies.

Given the long history of NLP and mental health studies, such as on social media \cite{9416889,DBLP:journals/corr/abs-2309-01618,DBLP:conf/icwsm/LokalaSD0APS22}, linguistic behavioral coding \cite{7362005}, and virtual counseling agents \cite{10015779,abs-2402-19052,DBLP:conf/kdd/SrivastavaSLA022}, there has been a significant leap in the conversational research space for multiple research applications. Presently, open-domain conversational systems such as ChatGPT \cite{openai2024gpt4}, XiaoIce \cite{xiaoice}, and GPT-3-based variants \cite{ZHANG2021831} excel in generating responses that are linguistically accurate and semantically coherent. 
However, in the sensitive context of counseling, these open-domain systems lack the capability to sufficiently meet the requirements \cite{10066740}.
On the other hand, certain approaches have focused on utilizing conventional methods to generate responses for mental health counseling \cite{NIPS2017_077e29b1, shen-etal-2017-conditional, wu-etal-2018-dialog}. More recent works have researched into controlling the response generation methodologies with factors like dialogue-acts \cite{10.1145/3543507.3583380}, empathy \cite{10.1145/3477495.3531912, 10.1145/3442381.3450097}, target \cite{tang-etal-2019-target}, and related aspects. These conventional methods were seen to be helpful only for a handful of specific cases, leaving room for dedicated research in the space of mental health counseling to tackle the nuanced demands of VMHAs. To address these gaps, we propose \model, a sentiment-guided commonsense-aware response generation mechanism for mental health counseling. We use a commonsense transformer to exploit commonsense knowledge and subsequently influence the client's sentiments in such a way that --  (i) if the client possesses a negative sentiment, \model's response influences the client toward the positive sentiment, and (ii) if the client is already in a positive state, \model's response should keep the client in the same sentiment state. This approach is motivated by the analysis of psychiatric conversations \cite{schueller2017ecological}, where clients express complex (mostly negative) sentiments, and the therapeutic process aims to guide them toward a more positive state \cite{sullivan2014strategies}. At last, we employ GPT-2 with modified knowledge-aware attention to learn commonsense and sentimental influence. As a result, the proposed system helps stabilize a client's sentimental state, making them more receptive toward subsequent treatment.

We evaluate \model\ using the HOPE dataset, a recognized benchmark in counseling \cite{malhotra2021speaker}, and compare its performance against $12$ potential baselines. Our approach demonstrates significant superiority over all the baselines across five key metrics. Furthermore, through empirical analysis and human evaluation, we ascertain that \model's generation capability is well-aligned with expectations, even surpassing the gold standard in certain instances. Additionally, we deploy \model\ as a chat interface for individuals seeking mental health support, evaluating its effectiveness through a comprehensive user study in a constrained setting. Findings indicate a notable positive response, with $91\%$ of users acknowledging the system's effectiveness, $80.25\%$ expressing satisfaction, and over $85\%$ expressing a willingness to continue using and recommending the interface, thereby underscoring the practical utility of \model. We summarize our contributions below\footnote{We plan to open-source the source code upon acceptance of the paper. We supplement the source-code for review here: https://shorturl.at/uyILM}:
\begin{itemize}[leftmargin=*]
    \item We propose \model, a novel transformer-based model for therapist response generation. It prioritizes the central objective of enhancing the therapeutic experience by influencing the client's sentiment toward the positive side.
    
    \item We exploit commonsense knowledge to understand clients better and construct sentiment-guided and commonsense-aware graph representations for counseling dialogue. 
    
    \item Our evaluation on the HOPE dataset shows \model's improved performance in generating therapist responses across $12$ competing baselines on five text-generation metrics.
    
    \item Our thorough qualitative analysis and human evaluation demonstrate \model's efficacy, which in some cases, surpasses the gold standard.

    \item We deploy \model-assistant as a chat interface for individuals seeking support. We conduct an exhaustive user study to evaluate the deployed system's effectiveness.
\end{itemize}

\noindent{\bf Reproducibility.} We open-source the proposed model at  for research purposes at \url{https://github.com/LCS2-IIITD/EmpRes}.

\section{Related Work}
\textbf{Commonsense Knowledge for Dialogue Generation.}
The positive outcomes of exploiting commonsense knowledge in AI and, more precisely, in NLP have been well studied \cite{ilievski2021dimensions}. Most of the early research primarily focused on incorporating commonsense knowledge in improving various aspects of text generation, including textual inference, story comprehension, and generation \cite{tandon2018commonsense, chaturvedi2017story,schlicht2021leveraging}. Commonsense is vital for generating coherent, contextually appropriate, and plausible text by providing background information and logical connections between ideas. As research advanced, there was a growing interest in conversational systems and domain-specific assistants, and therefore, there was a need for applying commonsense to dialogue generation for more relevant responses. \cite{holtzman2019curious} highlighted the problem of neural text degeneration, in which dialogue models tend to generate generic and uninformative responses. It supports the argument that there is a need to incorporate commonsense knowledge to avoid such degeneration. Recent advancements in dialogue generation leveraging commonsense knowledge are using commonsense transformer (COMET) \cite{DBLP:journals/corr/abs-1906-05317}. COMET is a generative model that possesses rich commonsense understanding by using deep pretrained language models. Using ATOMIC and Conceptnet, which are external knowledge graphs, authors capture different aspects of commonsense knowledge. Several studies \cite{shen2022knowledge,tu2022misc,li2022c3kg} highlighted the use of the COMET model in different contexts, such as counseling, Chinese dialogue and emotional support. These models seek to enhance dialogue responses by integrating external commonsense knowledge, but exploiting commonsense knowledge for text generation is still relatively nascent. Challenges include acquiring, representing, and effectively integrating external commonsense knowledge, making the generation process more complex.

\noindent\textbf{Response Generation.}
Response generation is a fundamental aspect of dialogue systems, in which the system generates appropriate and contextually relevant responses based on user input. Several techniques, along with commonsense to control response generation, are available. A few earlier works have highlighted the importance of incorporating emotion into response generation for empathetic and personalized dialogue systems \cite{chen2022emphi, gao2021improving, firdaus2021seprg}. By incorporating techniques such as selecting appropriate exemplars based on emotional content, recognizing emotional causes in conversations, and considering both user sentiment and desired emotional tone, they can use emotion in their response. Additional research works considered response generation as a sequential decision-making problem \cite{li2016deep, saha2020reinforcement}. This approach consists of a dialogue agent that learns to generate responses by interacting with an environment, receiving rewards based on the quality of its responses, and updating its policy through trial and error. By leveraging reinforcement learning algorithms, dialogue agents can optimize their response generation strategies to improve dialogue quality and achieve specific objectives, such as coherence, informativeness, or engaging conversations. On the other hand, there exist works that employ graph structure to capture crucial aspects of conversation context, including self-other awareness, causality reasoning, and multi-party dependencies, which account for participant relationships, causal dependencies, and diverse contextual information, leading to a contextually relevant and engaging dialogue \cite{wang2022care, gu2022hetermpc, zhao-etal-2019-rethinking}. This motivated us to design a sentiment-guided commonsense relation graph.

\section{Dataset}
We use HOPE \cite{malhotra2021speaker}, a benchmark counseling conversation dataset. It contains $12.8K$ utterances from $212$ dyadic counseling sessions between therapists and clients, publicly available on social platforms. The conversation in the dataset encompasses a wide range of demographic groups, each engaging in unique discussions related to mental health. To ensure data quality, earlier work was conducted to thoroughly process the transcriptions, eliminating any instances of noise or transcription errors \cite{malhotra2021speaker}. The dialogues collected for further analysis adopt a dyadic structure involving only two participants -- clients and therapists. To enhance the dataset, we apply a pseudo-labeling approach, tagging each utterance with its corresponding sentiment label. A detailed statistic of the HOPE dataset is presented in Table \ref{tab:counts}. We exploit the popular BERT-based fused with commonsense knowledge for the sentiment classification model to perform pseudo-labeling\footnote{nlptown/bert-base-multilingual-uncased-sentiment}. This auxiliary binary classification task enables us to tag $12.8K$ utterances with sentiment labels. In our analysis, we observe an uneven distribution of emotion labels (as it should be), particularly between positive and negative sentiments. While therapists predominantly maintain a neutral stance in most cases, the binary nature of emotion labels leads to a considerable presence of negative sentiment labels as well. We discuss the additional details about pseudo labeling in the next section.

% \begin{table}
% \centering

% \resizebox{0.485\textwidth}{!}
% {
% \begin{tabular}{l|cccc}
% \toprule
% HOPE & Train & Validation & Test & Total \\
% \cmidrule{1-5}
% Dialogue Sessions & 149 & 21 & 43 & 212\\
% Client Utterances  & 4668 & 595 & 1119 & 6382\\
% Therapist Utterances  & 4751 & 599 & 1122 & 6472\\
% \cdashline{1-5}
% \#Total Utterances & 9419 & 1194 & 2241 & 12854\\
% \bottomrule
% \end{tabular}
% }
% \caption{Statistics of the HOPE dataset \cite{malhotra2021speaker}. The dyadic counseling conversational dataset contains a total of 12.8k utterances, each associated with one of the 12 dialogue-act labels.}
% \label{tab:counts}
% \end{table}

\begin{table}
\centering
\resizebox{0.485\textwidth}{!}
{
\begin{tabular}{l|ccc}
\hline
 & Utterances & Positive & Negative \\
\hline
% Dialogue Sessions & 149 & 21 & 43 & 212 \\
Therapist Utterances  & 6472 & 4569 & 1903 \\
Client Utterances  & 6382 & 2540 & 3842 \\
\cdashline{1-4}
\#Total Utterances & 12854 & 7109 & 5745\\

\bottomrule
\end{tabular}
}
\caption{Statistics of the HOPE dataset \cite{malhotra2021speaker}. The dyadic counseling conversational dataset contains a total of $12.8k$ utterances, each tagged with positive/negative sentiment.}
\label{tab:counts}
 \vspace{-5mm}
\end{table}

% HOPE & Sp & Train & Validation & Test & Total \\
% \cmidrule{1-6}
% Dialogue Sessions & Th & 149 & 21 & 43 & 212\\
% \cmidrule{1-6}
% \multirow{2}*{Client Utterances} & Cl & 4668 & 595 & 1119 & 6382\\
%  & T/P & 4668 & 595 & 1119 & 6382\\
% \cmidrule{1-6}

\subsection{Dataset Extension}
\label{ap:dataset}
For our research, we utilized the HOPE benchmark counseling conversation dataset. This dataset comprises $212$ counseling conversations, amounting to approximately $\sim13$ utterances. To annotate the dataset with sentiment labels, we employ pseudo-labeling by utilizing both the commonsense transformer (COMET) and a BERT-based sentiment classification model. The central idea behind our approach is to utilize additional attributes derived from each utterance to enhance the performance of the sentiment classifier. Specifically, we used COMET to extract commonsense attributes related to the utterances, which we then integrated into the sentiment classification process. This integration aimed to provide the classifier with more contextual information, subsequently improving its accuracy. After conducting iterative experiments with various foundation models for sentiment classification, we observed that our approach performed marginally better than more sophisticated models. This led us to opt for a simpler and lighter foundation model, which balanced performance with efficiency. 

\begin{figure}[t]
  \centering
  \includegraphics[width=\columnwidth]{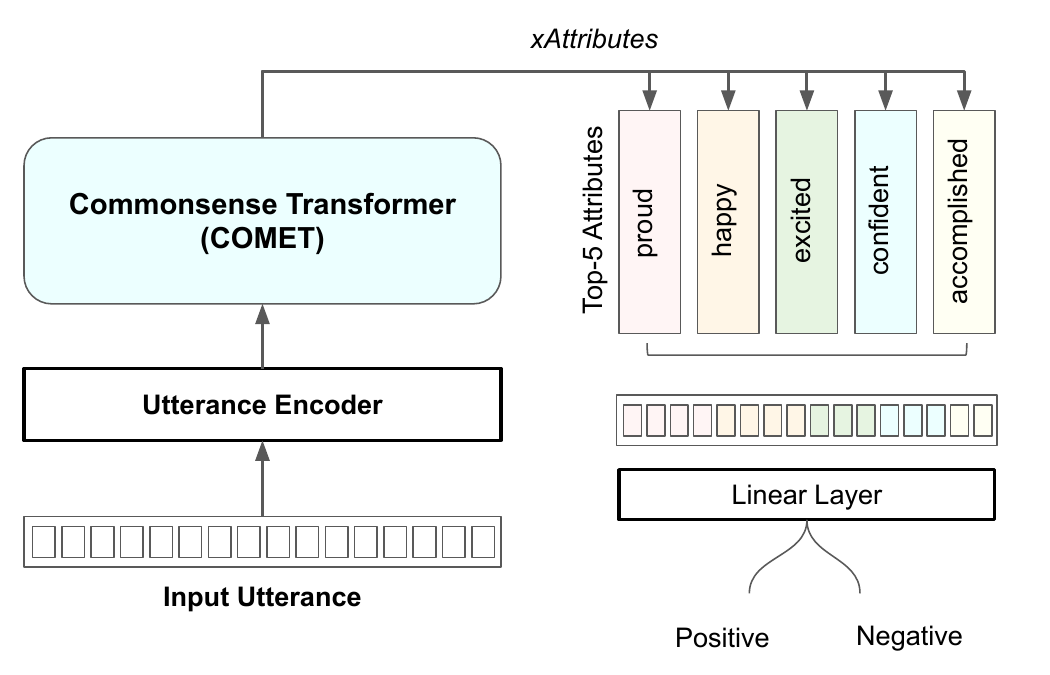}
  \caption{A schematic diagram of pseudo labeling for sentiment labels for each utterance in the HOPE dataset. We augment top-5 additional attributes using COMET to assist state-of-the-art sentiment classifier in predicting sentiment label.}
  \label{fig:sentimentlabel}
\end{figure}

The overall dataset extension pipeline is illustrated in Figure \ref{fig:sentimentlabel}. In this pipeline, the utterance encoder first processes the input utterance. Subsequently, COMET retrieves the {\em xAttribute} relations for the utterance, providing a set of attributes that describe likely properties or characteristics associated with the utterance. To ensure robustness, we selected the top-5 attributes from COMET for each utterance. These selected attributes were then concatenated with the original utterance using a {\em [SEP]} tag, creating an enriched input that was passed through a feed-forward layer for final classification.

\subsection{Discussion on COMET Relation Selection}
\label{ap:relation}

The selection of commonsense relations plays a crucial role in our model's understanding of the sentiment and intent of the clients. To assess the impact of different relations, we individually ran the commonsense transformer (COMET) on a sample of the HOPE dataset. For this purpose, we employ all nine COMET's relations -- {\em oEffect, oReact, oWant, xAttr, xEffect, xIntent, xNeed, xReact,} and {\em xWant}. We observe that the {\em xAttr} relation provides valuable insights into the specific attributes or characteristics related to the client's utterance. By analyzing this relation, we can determine whether the client is experiencing positive or negative emotions, which aids in sentiment classification. Hence, we employ such relation in the sentiment classification task as discussed in the Section \ref{ap:dataset}.

At a moment, a client can be found in one out of two possible states -- (i) positive or (ii) negative. In order to positively influence the client, we formulate the problem to fetch the commonsense knowledge in such a way that it conditions the knowledge extractor based on the client's last sentimental state. In general, if a client falls into a negative sentimental state, others (in our case, the therapist) consoles the client and influences them toward a positive side. Hence, we employ {\em oReact} and {\em oWant} relations effectively to capture what others would want for the client and how they feel about the client, respectively. These relations demonstrate a positive, helpful, and sentimental understanding of the client's negative sentiments. On the other hand, if a client feels positive, we intend to keep the client in the same state only. In addition, whatever client's thought is bringing positivity, we further exploit it in our response via three specific and meticulously analyzed relations -- {\em xReact, xWant, and xIntent}. These relations augment the self-feelings and self-thoughts, which further assists the VMHAs in generating responses capable of embedding similar thoughts and feelings into the response. 

Overall, the selection of commonsense relations plays a significant role in capturing the nuances of sentimental influence in our model. By leveraging the most relevant relations, we ensure a more accurate representation of clients' sentimental states and enable our model to provide supportive responses.

\begin{figure*}[ht]
  \centering
  \includegraphics[width=1.0\textwidth]{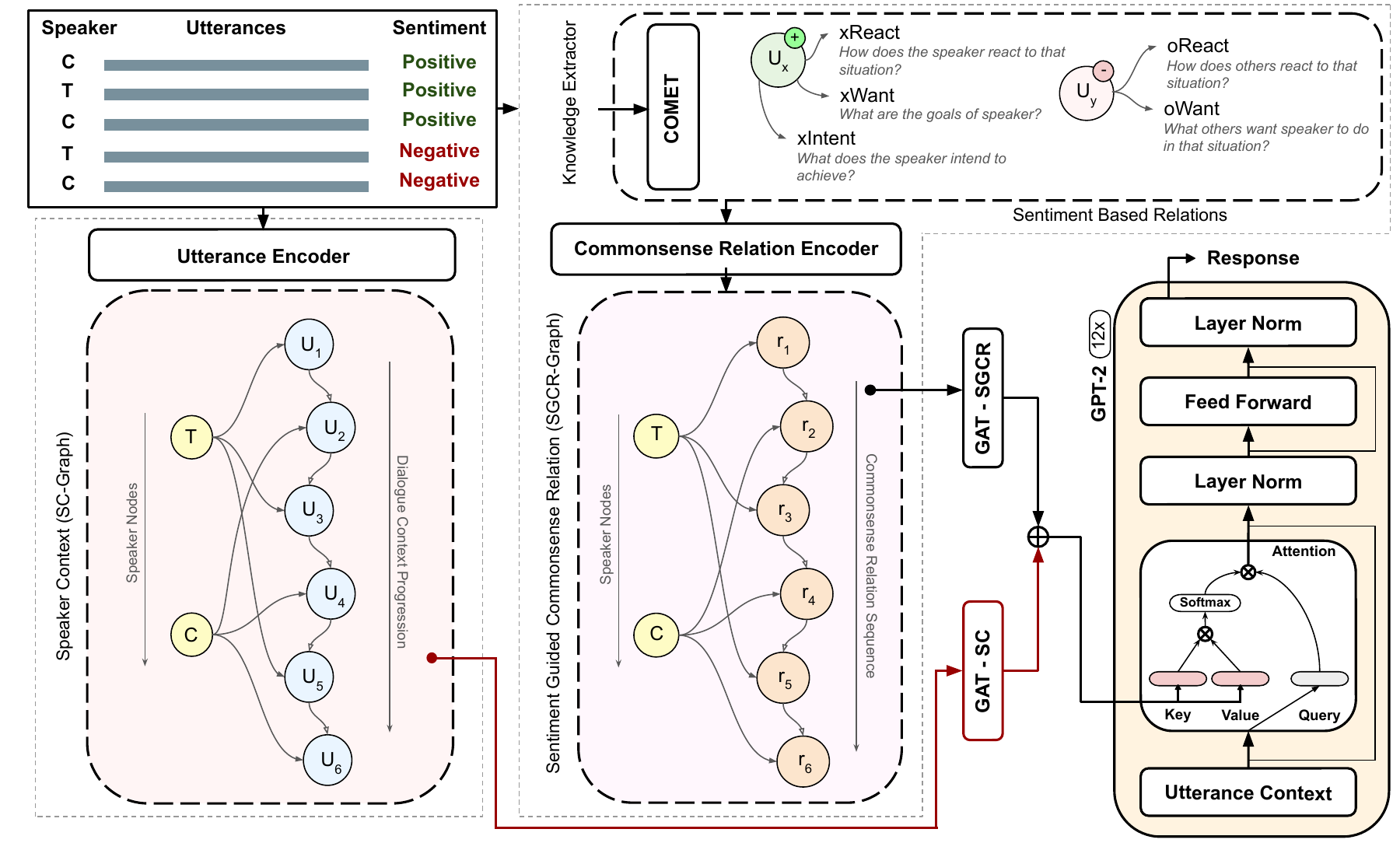}
  \caption{A schematic diagram of \model. {\em Knowledge Extractor} exploits conditional commonsense relations $(r_i)$ to construct Sentiment-Guided Commonsense Relation Graph (SGCR-Graph). A graph attention layer (GAT-SGCR), on top of SGCR-Graph, formulates sentiment-guided commonsense-aware representations. A similar layer on Speaker-Context Graph (SC-Graph), called GAT-SC, is responsible for formulating dialogue representations. Fused graph representations act as key and value for GPT-2's attention block, responsible for therapist response generation.}
  \label{fig:model-architecture}
\end{figure*}

\section{Proposed Methodology}
In a de facto setting, clients and virtual assistant exchange utterances. During this conversation, clients tend to express their sentiments, and VMHAs are expected to utilize commonsense to gain a clear understanding and subsequently generate responses. 

\noindent{\bf Problem Formulation.} Suppose there is a counseling dialogue between a therapist and a client containing utterances and their sentiment labels as $U \in \{u_0,u_1,\ldots, u_{t}\}$ and $S \in \{s_0,s_1,\ldots, s_{t}\}$ respectively, with $t$ being the time step. Here, sentiment labels are either \textit{positive} or \textit{negative} along speaker labels, $S_p \in \{ T, C \}$,  corresponding to each utterance. Our twofold jointly-learned tasks include  (i) exploiting COMET to generate sentiment-guided commonsense-aware dialogue representations (auxiliary), and (ii)  generating an effective therapist response $u_{t+1}$ against a client's utterance with an intent to keep the client with positive sentimental feeling (primary).

To this end, we propose \model, a novel commonsense aware sentiment-guided response generation mechanism for VMHAs. 
\model\ integrates the effectiveness of commonsense knowledge and sentiment guidance to place the utmost importance on fostering positive sentimental states in clients. Figure \ref{fig:model-architecture} illustrates the schematic framework of \model. 
\model\ generates the therapist's response against the client's last utterance. Initially, we construct a Speaker-Context Graph (SC-Graph) utilizing BERT to capture comprehensive representations of utterances and their relations with speaker information. On the other hand, the {\it Knowledge Extractor} module leverages a commonsense transformer (in our case, COMET) to extract relevant commonsense knowledge, taking into account the associated sentiment labels. To maintain a positive sentiment within the client, we conditionally employ distinct commonsense relations for utterances with positive and negative sentiments. This conceptual framework enables us to incorporate sentiment-guided knowledge effectively, thereby assisting in the construction of the Sentiment-Guided Commonsense Relation Graph (SGCR-Graph). A collective information from both SC-Graph and SGCR-Graph helps to train the foundation model (in our case, GPT-2) and subsequently generates responses that elicit positive sentiment labels throughout conversations. We provide details of each module below.

\subsection{Speaker Context Graph (SC-Graph)}
Dialogue structure is maintained by the utterance and the speaker's information. To effectively capture and maintain the dialogue structure, we utilize the Speaker Context Graph (SC-Graph). Such an approach models the dialogue by treating each utterance and its corresponding speaker information as nodes within the graph, thus encapsulating the relevant context for each part of the conversation. 

In the SC-Graph, each utterance node is connected via an edge to its respective speaker label, $SC$-$Graph(S_p \rightarrow u_t)$. This connection ensures that the speaker's identity is directly associated with their respective utterance, preserving the speaker-specific context. Furthermore, the sequential nature of the dialogue is maintained by connecting consecutive utterances through edges, represented as $SC$-$Graph(u_t \rightarrow u_{t+1})$. This structure captures the chronological progression of the conversation, ensuring that the temporal context between utterances is preserved.

To learn the contextual graph representations derived from the SC-Graph, we employ a graph attention layer (GAT-SC). This layer learns attention weaights allowing the model to assess the importance of different nodes and edges selectively and generate richer contextual representations.

\subsection{Commonsense Knowledge}
We employ a commonsense transformer, COMET, to extract sentiment-specific knowledge pertaining to each utterance. This extraction process involves the utilization of two distinct modules within it's pipeline -- (i) Knowledge Extractor, and further, this acquired knowledge is used to construct (ii) Sentiment-Guided Commonsense Relation Graph (SGCR-Graph). 

\noindent{\bf Knowledge Extractor.} 
For each utterance-sentiment tuple $\langle u_i, s_i \rangle$, we define a conditional commonsense knowledge extraction rule. We feed a pretrained commonsense transformer with $u_i$ as input by retrieving relevant knowledge aligned with sentiment-guided relations. For utterances expressing positive sentiments, we select COMET relations such as {\em xReact}, {\em xWant}, and {\em xIntent}. This selection is supported not only by empirical observations but also by psychological theories,  indicating that individuals tend to experience a psychologically happier state of mind when expressing positive sentiments and primarily focusing on themselves \cite{amutio2015mindfulness, karkar2021understanding}. Conversely, for utterances reflecting negative sentiments, we employ COMET relations such as {\em oReact} and {\em oWant}. In these cases, clients typically perceive themselves in a negative state of mind, prompting others (in our case, the therapist) to offer consolation and encourage a shift toward positivity. Considering the robustness of the model, we fetch top-5 knowledge inferences for each utterance. Once knowledge from the commonsense transformer is retrieved, we construct an SGCR-Graph associating the utterance-specific knowledge and speaker labels. We discussed the commonsense relation selection in detail in Section \ref{ap:relation}.

\noindent{\bf Sentiment Guided Commonsense Relation Graph (SGCR-Graph).}
The process begins with encoding the knowledge extracted by the knowledge extractor using {\em Knowledge-Encoder}. This module is responsible for generating meaningful representations of the input knowledge. The input to the Knowledge-Encoder consists of a concatenation of a sentiment label prompt and the top-5 outputs of extracted knowledge, separated by a special separator tag used by the BERT transformer. This concatenation ensures that both sentiment and relevant knowledge are considered simultaneously during encoding.

{\em Graph Structure}: The resultant representations ($r_t$) for each utterance ($u_i$) are then used to construct a new graph. In this graph, edges are established between the speaker labels (Counselor (T) or Client (C)) and the utterance knowledge representations. This structure allows the graph to encapsulate both the speaker's identity and the associated commonsense knowledge for each utterance. To maintain the context and capture the dialogue progression, directed edges are also added between the knowledge representation nodes of consecutive utterances to ensure that the temporal flow of the dialogue is preserved within the graph structure.

To further capture the nuances of sentiment knowledge and commonsense, we apply an additional graph attention layer (GAT-SGCR). This layer selectively focuses on important nodes and edges, refining the contextual representations.

\subsection{Response Generation Block}
For generating responses, we utilize models like GPT-2, which are pretrained on extensive text corpora and capable of producing semantically rich text. However, these models typically lack mechanisms for controlled and customized text generation. To address this, we employ GPT-2, a decoder-only transformer, and enhance its capabilities by incorporating external knowledge using a novel approach called {\em Knowledge Aware Attention}. The parametric size of GPT-2 (and others) is particularly suitable for our approach, as it allows for effective control of external knowledge without introducing significant noise. Larger models, while more powerful, tend to dilute the impact of external knowledge.

{\bf Knowledge Aware Attention.} 
To integrate external knowledge into the GPT-2 model, we modify its multi-head attention block (MHA-Block). This modifies the conventional attention mechanism into a new cross-attention block designed to incorporate enriched knowledge representations. In our approach, the {\em query} in the attention mechanism is derived from the hidden representation of the previous decoder layer, following the standard dot product attention method. However, the {\em key} and {\em value} components are sourced from enriched representations obtained through two graph attention layers: $GAT-SC$ and $GAT-SGCR$. These layers provide contextual and sentiment-guided commonsense information, respectively.

The representations from GAT-SC and GAT-SGCR are concatenated and then passed into the {\em Knowledge Aware Attention} block. This fusion of external knowledge with the internal mechanisms of GPT-2 improves the model's ability to generate responses that are not only semantically coherent but also contextually and sentimentally informed. By integrating these enriched representations, we improve the model's capacity to control text generation, leading to more relevant responses.

\section{Experiments and Results}
We perform numerous experiments to evaluate the performance of  \model, on the HOPE dataset by comparing it against twelve potential baselines. In this section, we provide comprehensive information about the HOPE dataset, the selection of baselines, and the metrics chosen for evaluation. 

\begin{table*}[t]\centering

\resizebox{\textwidth}{!}
{
\begin{tabular}{llcccccccccccc}

\toprule
& \multirow{2}{*}{} &\multicolumn{3}{c}{R1} &\multicolumn{3}{c}{R2} &\multicolumn{3}{c}{RL} &\multirow{2}{*}{BS} &\multirow{2}{*}{METEOR} \\
\cmidrule{3-11}
% \cline{2-11}
& & P & R & F1 & P & R & F1 & P & R & F1 & & \\ \midrule
\multirow{9}{*}{\rotatebox{90}{Baselines}} 
& HRED \cite{10.5555/3016387.3016435} & $7.86$&$6.29$&$5.73$&$0.65$&$0.61$&$0.39$ &$7.01$&$8.00$&$7.87$&$0.2421$	&$0.298$ \\

& VHCR \cite{park-etal-2018-hierarchical} & $7.28$&$5.66$&$5.28$&$0.44$&$0.04$&$0.41$& $8.2$&$8.96$&$8.32$&$0.5030$&$0.3266$ \\
& HRED w/ Sp-Utt. Enc. \cite{https://doi.org/10.48550/arxiv.1907.05599} & 8.26 & 7.55& 7.08& 	0.69& 	0.83& 	0.653	& 8.38& 	7.53& 	7.07& 	0.4169	& 0.0465\\
& GPT2 \cite{radford2019language} & 11.13 & 9.23 & 9.23 & 1.02 & 1.52 & 1.13 & 11.49 & 12.12 & 9.26 & 0.4727 & 0.3494\\
& ConKADI \cite{wu-etal-2020-diverse} & 8.43 & 6.34 & 6.38 & 1.24 &	0.98 & 0.98 & 9.48 & 9.13 & 7.83 & 0.2745 & 0.2588\\
& DialoGPT \cite{zhang-etal-2020-dialogpt} & 11.00 & 10.27	& 10.27	& 1.05 & 
 1.74 & 1.30 & 12.04 & 11.73 & 10.10	& 0.5819 & 0.3623\\
& CoMAE \cite{zheng-etal-2021-comae} &  0.15 & 0.27 & 0.16 & 0.01 & 0.03 & 0.01 & 0.10 & 0.20 & 0.11 & 0.2930 & 0.4030\\
& ProphetNet \cite{qi-etal-2021-prophetnet} &8.60 & 8.31 & 6.63 & 1.05 & 0.97 & 1.01 & 8.51 & 8.83 & 7.22 & 0.5422 & 0.3998\\
& DialogVED \cite{chen-etal-2022-dialogved} & 9.88 & 10.70 & 8.75 & 1.11 & 1.29 & 1.46 & 10.02 & 11.5 & 9.05 & 0.5898 & 0.3711\\
& EmpHi \cite{chen2022emphi} &  6.45 & 4.48 & 4.49 & 1.14 & 1.02 & 0.93 & 11.52 & 11.23 & 8.57 & 0.4658 & 0.2758\\
& CEM \cite{Sabour_Zheng_Huang_2022} & 8.25 & 5.58 & 6.78 & 0.75 & 0.91 & 0.81 & 9.54 & 8.59 & 7.07 & 0.3910 & 0.1232\\
& READER \cite{10.1145/3543507.3583380} & 10.59 & 10.85 & 10.85 & 0.58 & 1.23 & 0.71 & 10.85 & 9.32 & 9.46 & 0.7600 & 0.2103\\
\midrule

% \multirow{1}{*}{\rotatebox{90}{O}} 
% & \model\ w/ GPT2 &12.41  &43.91 &16.12 &3.70 &13.72 &4.98 &11.92 &41.02 &16.30 &0.7656 &0.2098 \\
% & \model\ -- RAC-Head & 12.64 & 41.48 & 15.78 & 3.60 & 11.83 & 4.58 & 12.3 & 38.64 & 15.90 & 0.7628 & 0.2039 \\
% & \rowcolor{blue!20}
% \textbf{Ablations} & & & & & & & & & & & \\

% & \rowcolor{blue!20} \bf \model &\textbf{13.23} &\textbf{13.33} &\textbf{12.93} &\textbf{1.82} &\textbf{1.99} &\textbf{1.92} &\textbf{13.23} &\textbf{13.5} &\textbf{11.80} &\textbf{0.8532} &\textbf{0.5984} \\
& \bf \model &\textbf{13.23} &\textbf{13.33} &\textbf{12.93} &\textbf{1.82} &\textbf{1.99} &\textbf{1.92} &\textbf{13.23} &\textbf{13.5} &\textbf{11.80} &\textbf{0.8532} &\textbf{0.5984} \\

\midrule
% \textbf{Reward Ablation} & & & & & & & & & & & \\
\multirow{3}{*}{\rotatebox{90}{Ablations}} 

& $\qquad$ -- {\em SC-Graph}  -- {\em SGCR-Graph}  & 11.13 & 9.23 & 9.23 & 1.02 & 1.52 & 1.13 & 11.49 & 12.12 & 9.26 & 0.4727 & 0.3494\\
& $\qquad$ -- {\em SC-Graph} &12.94 & 12.84 & 11.74 & 1.68 & 1.73 & 1.82 & 12.92 & 13.02 & 11.47 & 0.7654 & 0.5000\\
& $\qquad$ -- {\em SGCR-Graph} &12.12 & 12.83 & 12.83 & 1.80 & 1.52 & 1.54 & 12.54 & 12.83 & 10.73 & 0.6363 & 0.4240\\

% \midrule
% & $\Delta_{\model-BEST}(\%)$ & \textcolor{blue}{$\uparrow 18.86$} & \textcolor{blue}{$\uparrow 22.85$} & \textcolor{blue}{$\uparrow 19.17$} & \textcolor{blue}{$\uparrow 46.77$} & \textcolor{blue}{$\uparrow 14.36$} & \textcolor{blue}{$\uparrow 47.69$} & \textcolor{blue}{$\uparrow 9.88$} & \textcolor{blue}{$\uparrow 11.38$} & \textcolor{blue}{$\uparrow 16.83$} & \textcolor{blue}{$\uparrow 12.26$} & \textcolor{blue}{$\uparrow 51.28$} \\ 
\midrule
& $\Delta_{\model-BEST}(\%)$ & \textcolor{black}{$\uparrow 18.86$} & \textcolor{black}{$\uparrow 22.85$} & \textcolor{black}{$\uparrow 19.17$} & \textcolor{black}{$\uparrow 46.77$} & \textcolor{black}{$\uparrow 14.36$} & \textcolor{black}{$\uparrow 47.69$} & \textcolor{black}{$\uparrow 9.88$} & \textcolor{black}{$\uparrow 11.38$} & \textcolor{black}{$\uparrow 16.83$} & \textcolor{black}{$\uparrow 12.26$} & \textcolor{black}{$\uparrow 51.28$} \\

\bottomrule

\end{tabular}

}

\caption{The upper half shows the results obtained on the HOPE dataset. We show ROUGE (1, 2, L), BERTScore (BS), and METEOR to assess the performance of the proposed model, \model. The lower half of the table shows the ablation study in which we present individual contributions of two essential graph components in \model. Here, \textit{SC-Graph} represents speaker-context graph, and \textit{SGCR} denotes sentiment-guided commonsense relation graph. $\Delta_{\model-BEST}(\%)$ shows the percentage increment in the final model's metric when compared with the baselines.}
\label{tab: results}
 % \vspace{-4mm}
\end{table*}

\subsection{Comparable Baselines}
We compare \model\ with 12 competing baselines, focusing on systems that leverage dialogue context, commonsense, and empathy for response generation. Notably, the selected baselines cover all possible aspects of comparison. However, to the best of our knowledge, no other system has addressed the sentimental influence of any one interlocutor. In this study, we consider the following baselines as part of our analysis. 

\begin{table*}[]
    % \footnotesize
    \renewcommand*{\arraystretch}{1.0}
    \centering
    \resizebox{\textwidth}{!}{
    \begin{tabular}{p{32em}p{30em}}
        \toprule
        \bf {\centering Context} & \bf Therapist's Response \\ 
        \midrule 
        \multirow{2}{32em}{\textbf{Therapist:} How much are you drinking?} & 
        \multirow{2}{30em}{\textbf{Gold:} Yeah, are your friend's the problem? Maybe you just start hanging out with more kids in the youth group and not with your party kids?} \\ 
        % & & \\  
        & \\ 
        \cmidrule{2-2}
        \multirow{3}{32em}{{\bf Client:} I don't think I am drinking that much. I mean, it's just social gathering. Like its nothing that I do like by myself or whatever.} & 
        \multirow{3}{30em}{\textbf{\model:} {You haven't been able to have an easier time getting back into good habits.} {\em We can work together to identify strategies that will support your journey toward positive habits.} {\it [Positive Influence]}} \\ 
        & \\ 
        & \\ 
        \cmidrule{2-2} 
        \multirow{1}{32em}{\textbf{Therapist:} I am just so concerned. Can't you think of anything better to do?}  & 
        \multirow{1}{30em}{\textbf{GPT2:} You are going through something?} \\  
        % & & \\  
        \cmidrule{2-2}
        \multirow{2}{32em}{\textbf{Client:} I guess I can probably be more productive in my school work rather than going out to parties as much. But I feel helpless. {\bf {\em [Negative Sentiment]}}} & 
        \multirow{2}{30em}{\textbf{ChatGPT:}  Exploring healthier coping mechanisms and finding fulfilling activities can greatly improve your overall well-being.}\\ 
        & \\ 
        
        \midrule

        \multirow{1}{32em}{\textbf{Therapist:} You were in asylum for two weeks. How are you feeling now?} & 
        \multirow{1}{30em}{\textbf{Gold:} Tell me more about your new life.} \\ 
        % & & \\  
        \cmidrule{2-2}
        \multirow{2}{32em}{{\bf Client:} Yes, initially it was good then I got bored of daily stuff. I think, you know, it helped my craving for smokes though.} & 
        \multirow{2}{30em}{\textbf{\model:} {\em That is an incredible improvement.} Do you want to discuss about what other improvements you see?{\it [Positive Influence]}} \\ 
        & \\ 
        \cmidrule{2-2} 
        \multirow{2}{32em}{\textbf{Therapist:} Can you tell me if you have smoked after getting discharged?}  & 
        \multirow{2}{30em}{\textbf{GPT2:} It is hard to live with such habits and no self control.} \\  
        % & & \\  
        & \\ \cmidrule{2-2}
        \multirow{2}{32em}{\textbf{Client:} I think sometimes cravings can go high but my self control over my thoughts has improved so thanks to you. {\bf {\em [Positive Sentiment]}} } & 
        \multirow{2}{30em}{\textbf{ChatGPT:} Acknowledging cravings without acting on them is a crucial step. Just track your progress, and find healthier coping mechanisms.} \\ 
        & \\
        \midrule
        \multicolumn{2}{c}{\bf Error Analysis}\\
        
        \midrule

        \multirow{3}{32em}{\textbf{Therapist:} I think that's right. I see you crawled up over there. Also, you have a worried face and you're having hard time approaching it. You're feeling fear, but do not worry. When was the last time you had cheesecake?} & 
        \multirow{3}{30em}{\textbf{Gold:} I know, and so you've been avoiding food like this for so long that it totally makes sense that you're feeling fear.} \\ 
        & \\  
        & \\  
        \cmidrule{2-2}
        \multirow{1}{32em}{{\bf Client:} I honestly can't remember. I don't know.} & 
        \multirow{1}{30em}{\textbf{\model:} You are so much more than just being scared about this. } \\ 
        % & \\ 
        \cmidrule{2-2} 
        \multirow{2}{32em}{\textbf{Therapist:} I know it's been a really long time. You certainly didn't need it all through when you were really stuck in your eating disorder.}  & 
        \multirow{2}{30em}{\textbf{GPT2:} When was the last time you had cheesecake?} \\  
        % & & \\  
        & \\ \cmidrule{2-2}
        \multirow{2}{32em}{\textbf{Client:} I used to, I used to like it. I remember that, but it was a long time ago. Yet controlling is hitting hard on me. {\bf {\em [Negative Sentiment]}} } & 
        \multirow{2}{30em}{\textbf{ChatGPT:} It sounds cheesecake holds some nostalgic memory for you, but it's also connected to a difficult time with your eating disorder.} \\ 
        & \\
        \bottomrule
    \end{tabular}}
    \caption{A comparative analysis of therapist's response by the proposed model, \model, in contrast with the best-performing baseline (GPT-2), gold responses, and a state-of-the-art model, ChatGPT. Unlike other models, the responses generated by \model\ possess a positive influence on the client's negative sentiments. This is evident via the model's ability to control the response generation by exploiting {\em sentimental control} and {\em commonsense awareness}.}
    \label{tab:output}
    % \vspace{-2mm}
\end{table*}

\begin{itemize}
\item \textbf{HRED} is based on a hierarchical encoder-decoder and trained for the dialogue modeling \cite{10.5555/3016387.3016435}.
\item \textbf{VHCR} is based on variational hierarchical RNNs for dialogue modeling \cite{park-etal-2018-hierarchical}.
\item \textbf{HRED with Speaker and Utterance Encoder} enhances the HRED framework by incorporating speaker and utterance level information \cite{https://doi.org/10.48550/arxiv.1907.05599}. 
\item \textbf{GPT-2} is a decoder-only transformer model. Our model, \model, is also developed on GPT-2 \cite{radford2019language}.
\item \textbf{ConKADI} is an early effort in commonsense aware response generation \cite{wu-etal-2020-diverse}.
\item \textbf{DialoGPT} is a pretrained language model trained specifically on dialogue corpus \cite{zhang-etal-2020-dialogpt}. 
\item \textbf{CoMAE}  is a multi-factor hierarchical framework trained for empathetic response generation \cite{zheng-etal-2021-comae}.
\item \textbf{ProphetNet-Dialog} is trained to chat naturally and generate coherent responses \cite{qi-etal-2021-prophetnet}. 
\item \textbf{DialogVED} introduces continuous latent variables into the pretraining framework to enhance the relevance and diversity \cite{chen-etal-2022-dialogved}.
\item \textbf{EmpHi} focuses to generate empathetic dialogue responses \cite{chen2022emphi}.
\item \textbf{CEM} leverages the user's cognitive understanding for dialogue generation \cite{Sabour_Zheng_Huang_2022}.
\item \textbf{READER} is a transformer-reinforcement-learning approach for dialogue-act guided response generation. It is also trained on counseling conversational dataset \cite{10.1145/3543507.3583380}.
\end{itemize}

To assess the comparative performance of all baseline approaches and \model, we evaluate them using three widely-used evaluation metrics, namely ROUGE, METEOR, and BERTScore. The computation of these metrics is facilitated by specific libraries -- \textit{py-rouge}\footnote{https://pypi.org/project/py-rouge/} for ROUGE scores, \textit{nltk-meteor}\footnote{https://www.nltk.org/api/nltk.translate.meteor\_score.html} for METEOR, and \textit{Hugging Face - BERTScore}\footnote{https://huggingface.co/spaces/evaluate-metric/bertscore} for  BERTScore.

\subsection{Performance Comparison}
Table \ref{tab: results} shows the performance of the competing models. The results demonstrate the superiority of \model\ across all evaluation metrics. Notably, our model excels in capturing both semantic and syntactic structures, as evidenced by the significant improvements of +12.26\% and +51.28\% in BERTScore and METEOR metrics, respectively. Interestingly, none of the baselines were able to capture the $n$-gram overlap ($n\geq2$) efficiently, whereas \model\ receives a significant increase of 47.69\% in F1 and 46.77\% in  ROUGE-2 precision. In addition, \model\ consistently outperforms the baselines in ROUGE-1 with an improvement of 18.86\%, +22.85\%, and +19.17\% across precision, recall, and F1, respectively.

\subsection{Ablation Study}
\model\ comprises two crucial modules, namely the Speaker-Context Graph composed of ({\em SC-Graph}) and the Sentiment-Guided Commonsense Relation Graph and ({\em SGCR-Graph}), both of which are important in capturing the dialogue structure and fostering commonsense awareness. 

To thoroughly assess their impact, we conducted an ablation study as outlined in the latter portion of Table \ref{tab: results}. The results of this study underscore the importance of both components in enhancing \model's performance. Removing either the \textit{SC-Graph} or the \textit{SGCR-Graph} significantly influences the model's ability to generate coherent and contextually relevant responses. Interestingly, when the \textit{SC-Graph} is eliminated, leaving only the influence of the \textit{SGCR-Graph}, the model still demonstrates a commendable capability to produce responses that retain semantic richness and effectiveness.

This observation highlights the substantial contribution of sentimental influence and commonsense awareness to the dialogue generation process. Even though there is a marginal decrease in performance compared to the complete \model, the responses generated in the absence of the \textit{SC-Graph} remain remarkably close in quality. Thus, these findings reinforce the pivotal role played by sentimental cues and commonsense understanding in facilitating more natural and engaging conversational interactions.

\subsection{Qualitative Analysis}
To provide comprehensive support for our findings, we perform an extensive qualitative analysis. This analysis probes into the model's generation capability, followed by a detailed examination of the human evaluation scores. 

{\bf Generation Analysis.}
We investigate the quality of generations obtained from  \model. We observe that the responses generated by \model\ contain structured semantics as shown in Table \ref{tab:output}. To provide an estimate of the generative capabilities of other LLMs, we support \model's generations with gold responses from the HOPE dataset along with generations from GPT-2 and ChatGPT\footnote{https://chat.openai.com/}. Evidently, the \model\ is able to compete with the semantic generation standard of ChatGPT and surpasses the ability to influence the client in the counseling conversation positively. Table \ref{tab:output} highlights the section of \model's responses in which we observe sentimental control and commonsense awareness. We further observe that the responses received from \model\ are comparable to the gold responses and, in some cases, superior and influential.

{\bf Error Analysis.}
\model\ performs well in most of the cases. However, we uncover a few instances where a disproportionate emphasis on one of the integral components is observed. For instance, we scrutinize a counseling scenario in the {\em error analysis} section of Table \ref{tab:output}, which carries discussion related to eating disorders. Although \model's response adheres to the gold standard in terms of positive influence and sentiment, it falls short in terms of generating responses embedded with considerable knowledge. Our thorough analysis reveals that approximately 10\% of randomly sampled responses exhibit deficiencies in either sentimental influence or the incorporation of relevant commonsense knowledge. Not to ignore that such generations are also possible due to the weak gold standard as in many cases, our human evaluation further claims that there exists a significant number of generations surpassing the quality of the gold standard. These evaluations are discussed next.

\begin{table}[!t]
\centering
\resizebox{0.485\textwidth}{!}{
\begin{tabular}{l|cccc}
\toprule
Model & Relevance & Consistency & Fluency & Coherence \\
\cmidrule{1-5}
ChatGPT  & 2.13 & 1.89 & \bf 3.41 & 3.15\\
GPT2  & 2.08 & 2.46 & 2.27 & 3.06\\
\cdashline{1-5}
\textbf{\model} & \bf 3.18 ($\rho$=0.0048) & \bf 2.79 ($\rho$=0.0032) & 3.36 ($\rho$=0.0044) & \bf 3.21 ($\rho$=0.0052)\\
\bottomrule
\end{tabular}
}
\caption{Human evaluation on the responses generated by \model\  when compared to the best-performing baseline and ChatGPT, along with their statistical significance test (p-value, $\rho$). The performance of \model\ across all metrics is up to the mark and slightly better than comparing systems.}
\label{tab:humaneval}
\end{table}

{\bf Human Evaluation.}
To ensure the quality and effectiveness of  \model, we conduct a comprehensive human evaluation on linguistic grounds \cite{VANDERLEE2021101151}. We employ four dedicated metrics: 

\begin{itemize}
    \item \textbf{Relevance} shows the selection of relevant content in the generated response considering the reference utterance.
    \item \textbf{Consistency} evaluates the factual alignment between the generated response and the source utterance. 
    \item \textbf{Fluency} demonstrates the linguistic quality of the generated responses. 
    \item \textbf{Coherence} shows the structure and organization of the generated responses.
\end{itemize}

We conducted an assessment of $70$ randomly chosen instances, with $15$ linguistic evaluators aged between $25$ and $45$. Each evaluator was tasked with assigning a score between $1$ and $5$ to four distinct parameters, where $5$ represents the highest quality. This evaluation was repeated for both ChatGPT and GPT-2 for comparison purposes. By calculating the average scores obtained, we present the results in Table \ref{tab:humaneval}. We observe that \model\ consistently demonstrates its superior qualitative performance except {\em fluency}, where ChatGPT has surpassed due to its exceptional ability to formulate human-level generation. Evaluators further stated that the model, in some cases, is able to generate better than the gold standard.

{\bf Relevance to Mental Health Experts.} 
To further validate the work's relevance, we collaborated with a licensed clinical psychologist and professor from a prominent organization. The effectiveness of \model\ is validated through an exhaustive expert evaluation using three parameters on the scale of $1$ (absent) to $5$ (present).

\begin{itemize}
    \item {\bf Positive Impact} shows responses' positive contribution to the user's sentimental well-being.
    \item {\bf Commonsense Understanding} shows the incorporation of contextually relevant responses that reflect a commonsense from context.
    \item {\bf Therapeutic Appropriateness} ensures advice and guidance align with a therapeutic approach. 
\end{itemize}

We observe an average rating of 2.70, 2.80, and 2.60 on the \model's responses for positive impact, commonsense understanding, and therapeutic appropriateness parameters, respectively. In addition, experts have confirmed that our model's average rating surpasses that of competitive responses, further affirming the model's capabilities. On the other hand, experts also claim that the proposed system sometimes fails to reflect certain crucial elements, showing a scope for improvement for future studies.

\begin{figure*}[t]
  \centering
  \includegraphics[width=1.0\textwidth]{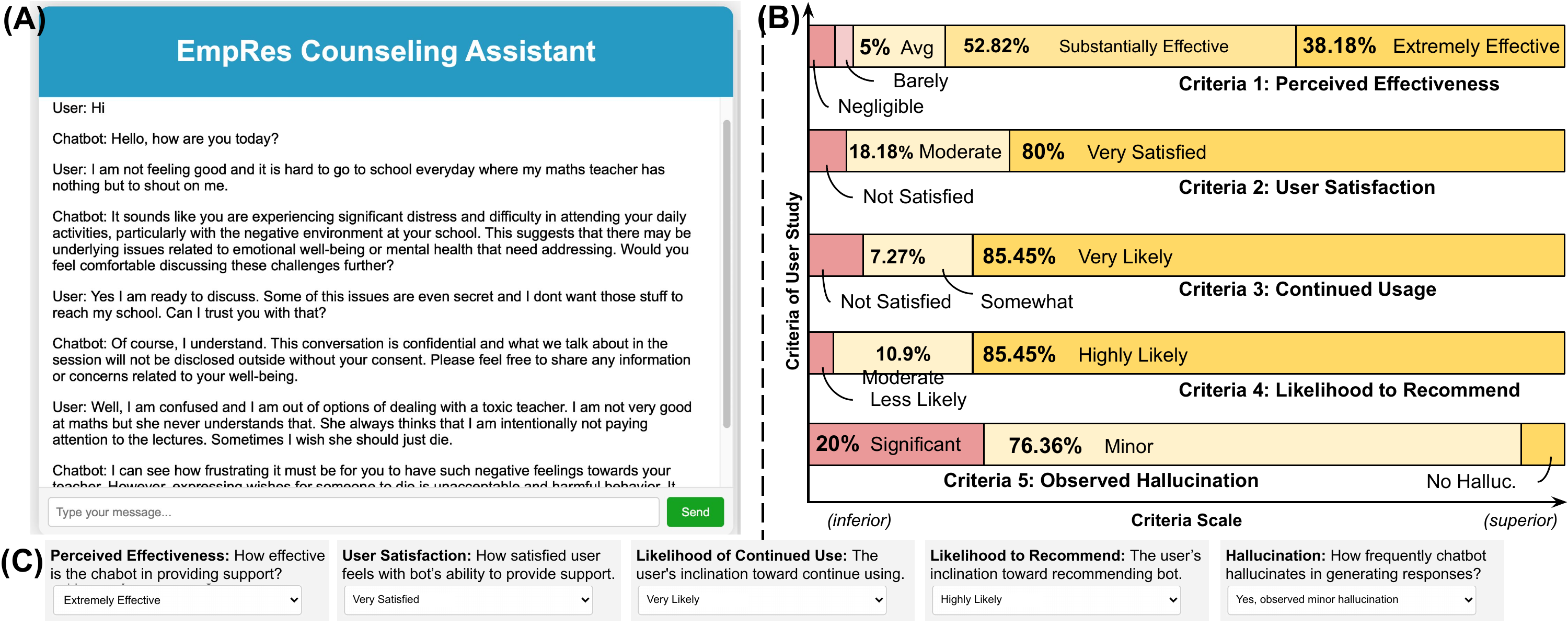}
  \caption{The figure illustrates (A) the prototype deployment of the chat interface and (B) findings of the user study conducted to evaluate the real-world effectiveness of the deployed \model. (C) The user study comprises user feedback of conversation with the deployed system across five major criteria -- (i) perceived effectiveness, (ii) user satisfaction, (iii) continued usage, (iv) likelihood to recommend, and (v) observed hallucination.}
  \label{fig:userstudy}
\end{figure*}

\subsection{Experimental Setup}\label{ap:experiment}
We perform numerous experiments using various combinations of the 
relevant commonsense relations, modeling loss. Further, we extensively experimented with graph construction. The process of graph construction for each batch is executed beforehand to keep the learning efficient. With this process, we significantly dropped the training time, making the model 10x more efficient. Moreover, we conduct extensive hyper-parameter tuning to accurately learn the joint learning framework and observe a significant boost in the BERTScore and METEOR metrics. 

We perform all experiments on an Nvidia A6000 GPU. We tune our hyperparameters to find the optimal configurations. We utilize the learning rate of $2\times10^{-6}$, batch size of $8$, which we run for $20$ epochs. We observe that the model provides a significant decrease in the loss for the first ten epochs and then stabilizes itself in the next ten epochs.

\section{Production Deployment}
We have implemented \model\ in a controlled yet natural setting to assess its efficacy, allowing users to interact with the assistant. The primary objective is to introduce this system to a group of users, providing them access to the interface and encouraging engagement with the assistant. This deployment serves as the foundation for a comprehensive user study aimed at evaluating the real-world effectiveness of \model. After gathering initial feedback from this study (stage 0), we advanced to stage 1, conducting a pilot study under expert supervision to explore usability aspects more thoroughly. The insights obtained from this pilot study will inform refinements to \model, enhancing its functionality in preparation for wider public usage in stage 2. Next, we present the findings from the initial user study conducted during stage 0.

\subsection{User Study}
We conducted a comprehensive user study to evaluate the real-world effectiveness and user satisfaction of \model\ as a well-being assistant. A diverse group of participants ($N=55$) were invited to engage with the assistant within a controlled setting and provide feedback on various interaction aspects. Invited participants represented a wide demographic range, including individuals from varying age groups (20-40), backgrounds, and geographic locations, to ensure a fair study and feedback mechanism. After each interaction, participants were prompted to provide feedback through a structured survey. 
The study focused on several key research questions to assess \model's performance pertaining to the perceived effectiveness, satisfaction, likelihood of continued usage, recommendation to others, and user's perceptions regarding hallucinations. Based on these aspects, we discuss the findings of the study. Figure \ref{fig:userstudy} summarises the key findings from the user study.

Firstly, regarding effectiveness and satisfaction, $91\%$ of users rated \model\ as {\em significantly or substantially effective} in providing mental health support. This high effectiveness rating indicates that \model\ has the potential to positively impact users' well-being. Additionally, $80\%$ of participants expressed satisfaction with their interactions, demonstrating a generally positive user experience.

Evidently, the results indicate a significant positive response, with $91\%$ of users rating \model\ to be {\em significantly} or {\em substantially effective} in providing mental health support. Additionally, $80\%$ expressed satisfaction, and over $85.45\%$ conveyed their likelihood of continued usage and recommendation to others. On the other hand, a minute portion of $20\%$ of users observed potential hallucinations in the model's responses, highlighting the importance of controlling the model's abilities to minimize such occurrences in future research.

\subsection{Ethical Considerations for User Study} 
While deploying \model\ as an assistant, we adhere to stringent ethical standards to safeguard user privacy, safety, and overall well-being. Prior to engaging with the assistant, we present participants in the user study with comprehensive guidelines and declarations. These declarations ensure that no personally identifiable information is collected within the deployed system, granting users the freedom to express themselves anonymously. In addition, participation in the study is entirely voluntary. The system adheres to ethical standards, emphasizing transparency and cultivating a positive user experience.

\subsection{Application Scenario} \model\ holds diverse and promising applications beyond traditional counselor assistance, extending its reach to become an invaluable tool for researchers in this field. Its potential applications are expansive. This shows promise in easing the counselor's process in various environments, including corporate wellness programs, employee assistance initiatives, and educational institutions. Looking beyond practical support roles, \model\ emerges as an innovative assistive system for researchers. It can be leveraged to understand counselor response behavior during counseling setups, providing a unique perspective and valuable insights for advancing mental health research. This section explores the broad spectrum of possibilities for \model, showcasing its potential impact across various applications.

\subsection{Deployment Limitations and Scalability}
While \model\ demonstrates remarkable effectiveness in providing support, it is essential to discuss certain constraints for its deployment and scalability in broader use cases. During the user study, we observe that there are minor instances of hallucination in \model's responses. Approximately $76\%$ of participants reported occasional {\em minor hallucination}, highlighting the need for improvement in our control over model's generation. On the other hand, the scalability of \model\ depends on the availability of counseling-relevant datasets. The success of \model\ hinges on the quality and diversity of the data it is trained on. To scale \model\ for larger use cases, there is a need for more comprehensive datasets specifically tailored to counseling contexts. Despite these limitations, the feedback received so far from the user study is overwhelmingly positive, underscoring the practical applicability of \model\ as an assistant. With ongoing refinement in training data, there is immense potential for \model\ to be scaled for broader use cases in stage 2, contributing to the advancement of systems on a larger scale.

\subsection{Societal Impact}
Our proposed method serves as a valuable contribution to the mental health community and ongoing scientific investigations by harnessing the advancements in current-age dialogue systems specifically meant for counseling purposes. Such novel contributions within the realm of mental health hold significant potential for delivering substantial societal benefits. By directly aiding mental health professionals and industries in this space and offering accessible assistance to individuals in need, our work aims to foster a positive social impact in this domain.

\section{Ethics and Limitations}
The research conducted in this sensitive area necessitates a comprehensive consideration of ethical implications and limitations. Given the gravity of the research, we prioritize the preservation of client privacy by implementing rigorous safeguards at every stage of our study. The study is backed with promising quantitative and qualitative results along with human feedback, who were adequately compensated. While the significance of this work to society cannot be overstated, it is important to acknowledge the scope of improvement in certain aspects, such as the model's ability to handle extreme cases such as suicidal thoughts and accommodate linguistic diversity such as languages other than English. 
Furthermore, it is crucial to recognize that our approach represents only one facet of the broader efforts in this space. To illustrate this further, consider the typical trajectory of counseling. In the initial stages, clients often experience emotional distress, making it challenging for them to assess their situation or engage in formal treatment. Our primary goal is to develop a system that can effectively stabilize a client's mental state, thereby making them more receptive to subsequent treatment. Therefore, its primary purpose is not posed to be an end-to-end treatment framework; instead, it serves as a crucial preliminary module within an end-to-end VMHA framework. 

\section{Conclusion}
The perpetual challenge of addressing the scarcity of mental health professionals has become increasingly pronounced with each passing year. While the emergence of Virtual Mental Health Assistants (VMHAs) has provided clients with convenient access to support, the generation of commonsense-grounded effective responses remains an underexplored area. In this paper, we presented \model, a novel model that leverages two fundamental graphs: the speaker-context graph and the sentiment-guided commonsense relation graph—to preserve dialogue structure and facilitate the extraction of relevant knowledge. Additionally, we employed GPT-2 with a modified knowledge-aware attention mechanism to incorporate augmented commonsense knowledge and sentimental influence into the response generation process. Our comprehensive evaluation against $12$ competitive baselines demonstrated the superiority of \model. Furthermore, we conducted an extensive ablation study, qualitative analysis, and human evaluation to highlight the effectiveness of \model and identify its limitations. To assess the real-world applicability of our system, we deployed \model as a chat interface and conducted a user study to evaluate its efficacy. The findings from the user study were promising, showing that $91.0\%$ users found the assistant's responses to be {\em effective}. On the other hand, $80\%$ users reported increased satisfaction with the interaction quality, noting that the responses felt more natural and supportive compared to other VMHAs they had previously used. The positive results from this deployment have laid the groundwork for further scalability and future enhancements. Our findings underscore the potential of \model to significantly improve the quality and responsiveness of VMHAs.

\bibliographystyle{IEEEtran}
\bibliography{sample-base}

\begin{thebibliography}{10}
\providecommand{\url}[1]{#1}
\csname url@samestyle\endcsname
\providecommand{\newblock}{\relax}
\providecommand{\bibinfo}[2]{#2}
\providecommand{\BIBentrySTDinterwordspacing}{\spaceskip=0pt\relax}
\providecommand{\BIBentryALTinterwordstretchfactor}{4}
\providecommand{\BIBentryALTinterwordspacing}{\spaceskip=\fontdimen2\font plus
\BIBentryALTinterwordstretchfactor\fontdimen3\font minus \fontdimen4\font\relax}
\providecommand{\BIBforeignlanguage}[2]{{%
\expandafter\ifx\csname l@#1\endcsname\relax
\typeout{** WARNING: IEEEtran.bst: No hyphenation pattern has been}%
\typeout{** loaded for the language `#1'. Using the pattern for}%
\typeout{** the default language instead.}%
\else
\language=\csname l@#1\endcsname
\fi
#2}}
\providecommand{\BIBdecl}{\relax}
\BIBdecl

\bibitem{deangelis}
\BIBentryALTinterwordspacing
T.~DeAngelis, ``Better relationships with patients lead to better outcomes.'' in \emph{Monitor on Psychology}.\hskip 1em plus 0.5em minus 0.4em\relax Monitor on Psychology, 2019. [Online]. Available: \url{https://www.apa.org/monitor/2019/11/ce-corner-relationships}
\BIBentrySTDinterwordspacing

\bibitem{9416889}
M.~E. Aragón, A.~P. López-Monroy, L.~C. González-Gurrola, and M.~Montes-y Gómez, ``Detecting mental disorders in social media through emotional patterns - the case of anorexia and depression,'' \emph{IEEE Transactions on Affective Computing}, vol.~14, no.~1, pp. 211--222, 2023.

\bibitem{DBLP:journals/corr/abs-2309-01618}
\BIBentryALTinterwordspacing
A.~Srivastava, T.~Gupta, A.~Cerezo, S.~P. Lord, M.~S. Akhtar, and T.~Chakraborty, ``Critical behavioral traits foster peer engagement in online mental health communities,'' \emph{CoRR}, vol. abs/2309.01618, 2023. [Online]. Available: \url{https://doi.org/10.48550/arXiv.2309.01618}
\BIBentrySTDinterwordspacing

\bibitem{DBLP:conf/icwsm/LokalaSD0APS22}
\BIBentryALTinterwordspacing
U.~Lokala, A.~Srivastava, T.~G. Dastidar, T.~Chakraborty, M.~S. Akhtar, M.~Panahiazar, and A.~P. Sheth, ``A computational approach to understand mental health from reddit: Knowledge-aware multitask learning framework,'' in \emph{Proceedings of the Sixteenth International {AAAI} Conference on Web and Social Media, {ICWSM} 2022, Atlanta, Georgia, USA, June 6-9, 2022}, C.~Budak, M.~Cha, and D.~Quercia, Eds.\hskip 1em plus 0.5em minus 0.4em\relax {AAAI} Press, 2022, pp. 640--650. [Online]. Available: \url{https://ojs.aaai.org/index.php/ICWSM/article/view/19322}
\BIBentrySTDinterwordspacing

\bibitem{7362005}
J.~Gibson, A.~Katsamanis, F.~Romero, B.~Xiao, P.~Georgiou, and S.~Narayanan, ``Multiple instance learning for behavioral coding,'' \emph{IEEE Transactions on Affective Computing}, vol.~8, no.~1, pp. 81--94, 2017.

\bibitem{10015779}
I.~Chatterjee, M.~Gorsic, M.~S. Hossain, J.~D. Clapp, and V.~D. Novak, ``Automated classification of dyadic conversation scenarios using autonomic nervous system responses,'' \emph{IEEE Transactions on Affective Computing}, vol.~14, no.~04, pp. 3388--3395, oct 2023.

\bibitem{abs-2402-19052}
\BIBentryALTinterwordspacing
P.~K. Adhikary, A.~Srivastava, S.~Kumar, S.~M. Singh, P.~Manuja, J.~K. Gopinath, V.~Krishnan, S.~Kedia, K.~S. Deb, and T.~Chakraborty, ``Exploring the efficacy of large language models in summarizing mental health counseling sessions: {A} benchmark study,'' \emph{CoRR}, vol. abs/2402.19052, 2024. [Online]. Available: \url{https://doi.org/10.48550/arXiv.2402.19052}
\BIBentrySTDinterwordspacing

\bibitem{DBLP:conf/kdd/SrivastavaSLA022}
\BIBentryALTinterwordspacing
A.~Srivastava, T.~Suresh, S.~P. Lord, M.~S. Akhtar, and T.~Chakraborty, ``Counseling summarization using mental health knowledge guided utterance filtering,'' in \emph{{KDD} '22: The 28th {ACM} {SIGKDD} Conference on Knowledge Discovery and Data Mining, Washington, DC, USA, August 14 - 18, 2022}, A.~Zhang and H.~Rangwala, Eds.\hskip 1em plus 0.5em minus 0.4em\relax {ACM}, 2022, pp. 3920--3930. [Online]. Available: \url{https://doi.org/10.1145/3534678.3539187}
\BIBentrySTDinterwordspacing

\bibitem{openai2024gpt4}
O.~Team, ``Gpt-4 technical report,'' 2024.

\bibitem{xiaoice}
L.~Zhou, J.~Gao, D.~Li, and H.-Y. Shum, ``The design and implementation of xiaoice, an empathetic social chatbot,'' \emph{Computational Linguistics}, vol.~46, pp. 1--62, 01 2020.

\bibitem{ZHANG2021831}
M.~Zhang and J.~Li, ``A commentary of gpt-3 in mit technology review 2021,'' \emph{Fundamental Research}, vol.~1, no.~6, pp. 831--833, 2021.

\bibitem{10066740}
L.~Brocki, G.~C. Dyer, A.~Gładka, and N.~C. Chung, ``Deep learning mental health dialogue system,'' in \emph{2023 IEEE International Conference on Big Data and Smart Computing (BigComp)}, 2023, pp. 395--398.

\bibitem{NIPS2017_077e29b1}
J.~Lu, A.~Kannan, J.~Yang, D.~Parikh, and D.~Batra, ``Best of both worlds: Transferring knowledge from discriminative learning to a generative visual dialog model,'' in \emph{NeurIPS}, I.~Guyon, U.~V. Luxburg, S.~Bengio, H.~Wallach, R.~Fergus, S.~Vishwanathan, and R.~Garnett, Eds., vol.~30.\hskip 1em plus 0.5em minus 0.4em\relax Curran Associates, Inc., 2017.

\bibitem{shen-etal-2017-conditional}
X.~Shen, H.~Su, Y.~Li, W.~Li, S.~Niu, Y.~Zhao, A.~Aizawa, and G.~Long, ``A conditional variational framework for dialog generation,'' in \emph{Proceedings of the 55th Annual Meeting of the ACL}.\hskip 1em plus 0.5em minus 0.4em\relax Vancouver, Canada: ACL, Jul. 2017, pp. 504--509.

\bibitem{wu-etal-2018-dialog}
X.~Wu, A.~Mart{\'\i}nez, and M.~Klyen, ``Dialog generation using multi-turn reasoning neural networks,'' in \emph{Proceedings of the 2018 Conference of the NAACL}.\hskip 1em plus 0.5em minus 0.4em\relax New Orleans, Louisiana: ACL, Jun. 2018, pp. 2049--2059.

\bibitem{10.1145/3543507.3583380}
A.~Srivastava, I.~Pandey, M.~S. Akhtar, and T.~Chakraborty, ``Response-act guided reinforced dialogue generation for mental health counseling,'' in \emph{Proceedings of the ACM Web Conference 2023}, ser. WWW '23.\hskip 1em plus 0.5em minus 0.4em\relax New York, NY, USA: Association for Computing Machinery, 2023, p. 1118–1129.

\bibitem{10.1145/3477495.3531912}
\BIBentryALTinterwordspacing
T.~Saha, V.~Gakhreja, A.~S. Das, S.~Chakraborty, and S.~Saha, ``Towards motivational and empathetic response generation in online mental health support,'' in \emph{Proceedings of the 45th International ACM SIGIR Conference on Research and Development in Information Retrieval}, ser. SIGIR '22.\hskip 1em plus 0.5em minus 0.4em\relax New York, NY, USA: ACM, 2022, p. 2650–2656. [Online]. Available: \url{https://doi.org/10.1145/3477495.3531912}
\BIBentrySTDinterwordspacing

\bibitem{10.1145/3442381.3450097}
\BIBentryALTinterwordspacing
A.~Sharma, I.~W. Lin, A.~S. Miner, D.~C. Atkins, and T.~Althoff, ``Towards facilitating empathic conversations in online mental health support: A reinforcement learning approach,'' in \emph{Proceedings of the Web Conference 2021}, ser. WWW '21.\hskip 1em plus 0.5em minus 0.4em\relax New York, NY, USA: ACM, 2021, p. 194–205. [Online]. Available: \url{https://doi.org/10.1145/3442381.3450097}
\BIBentrySTDinterwordspacing

\bibitem{tang-etal-2019-target}
J.~Tang, T.~Zhao, C.~Xiong, X.~Liang, E.~Xing, and Z.~Hu, ``Target-guided open-domain conversation,'' in \emph{Proceedings of the 57th Annual Meeting of the ACL}.\hskip 1em plus 0.5em minus 0.4em\relax Florence, Italy: ACL, Jul. 2019, pp. 5624--5634.

\bibitem{schueller2017ecological}
S.~M. Schueller, A.~Aguilera, and D.~C. Mohr, ``Ecological momentary interventions for depression and anxiety,'' \emph{Depression and anxiety}, vol.~34, no.~6, pp. 540--545, 2017.

\bibitem{sullivan2014strategies}
M.~J. Sullivan, ``Strategies for working with difficult clients,'' in \emph{Parenting coordination in post-separation disputes: A comprehensive guide for practitioners}, S.~A. Higuchi and S.~J. Lally, Eds.\hskip 1em plus 0.5em minus 0.4em\relax American Psychological Association, 2014, pp. 107--122.

\bibitem{malhotra2021speaker}
\BIBentryALTinterwordspacing
G.~Malhotra, A.~Waheed, A.~Srivastava, M.~S. Akhtar, and T.~Chakraborty, ``Speaker and time-aware joint contextual learning for dialogue-act classification in counselling conversations,'' in \emph{Proceedings of the Fifteenth ACM International Conference on Web Search and Data Mining}, ser. WSDM '22.\hskip 1em plus 0.5em minus 0.4em\relax New York, NY, USA: Association for Computing Machinery, 2022, p. 735–745. [Online]. Available: \url{https://doi.org/10.1145/3488560.3498509}
\BIBentrySTDinterwordspacing

\bibitem{ilievski2021dimensions}
F.~Ilievski, A.~Oltramari, K.~Ma, B.~Zhang, D.~L. McGuinness, and P.~Szekely, ``Dimensions of commonsense knowledge,'' \emph{Knowledge-Based Systems}, vol. 229, p. 107347, 2021.

\bibitem{tandon2018commonsense}
N.~Tandon, A.~S. Varde, and G.~de~Melo, ``Commonsense knowledge in machine intelligence,'' \emph{ACM SIGMOD Record}, vol.~46, no.~4, pp. 49--52, 2018.

\bibitem{chaturvedi2017story}
S.~Chaturvedi, H.~Peng, and D.~Roth, ``Story comprehension for predicting what happens next,'' in \emph{Proceedings of the 2017 Conference on EMNLP}, 2017, pp. 1603--1614.

\bibitem{schlicht2021leveraging}
I.~B. Schlicht, E.~Sezerer, S.~Tekir, O.~Han, and Z.~Boukhers, ``Leveraging commonsense knowledge on classifying false news and determining checkworthiness of claims,'' \emph{arXiv preprint arXiv:2108.03731}, 2021.

\bibitem{holtzman2019curious}
A.~Holtzman, J.~Buys, L.~Du, M.~Forbes, and Y.~Choi, ``The curious case of neural text degeneration,'' \emph{arXiv preprint arXiv:1904.09751}, 2019.

\bibitem{DBLP:journals/corr/abs-1906-05317}
\BIBentryALTinterwordspacing
A.~Bosselut, H.~Rashkin, M.~Sap, C.~Malaviya, A.~Celikyilmaz, and Y.~Choi, ``{COMET:} commonsense transformers for automatic knowledge graph construction,'' \emph{CoRR}, vol. abs/1906.05317, 2019. [Online]. Available: \url{http://arxiv.org/abs/1906.05317}
\BIBentrySTDinterwordspacing

\bibitem{shen2022knowledge}
S.~Shen, V.~P{\'e}rez-Rosas, C.~Welch, S.~Poria, and R.~Mihalcea, ``Knowledge enhanced reflection generation for counseling dialogues,'' in \emph{Proceedings of the 60th Annual Meeting of the Association for Computational Linguistics (Volume 1: Long Papers)}, 2022, pp. 3096--3107.

\bibitem{tu2022misc}
Q.~Tu, Y.~Li, J.~Cui, B.~Wang, J.-R. Wen, and R.~Yan, ``Misc: A mixed strategy-aware model integrating comet for emotional support conversation,'' \emph{arXiv preprint arXiv:2203.13560}, 2022.

\bibitem{li2022c3kg}
D.~Li, Y.~Li, J.~Zhang, K.~Li, C.~Wei, J.~Cui, and B.~Wang, ``C3kg: A chinese commonsense conversation knowledge graph,'' \emph{arXiv preprint arXiv:2204.02549}, 2022.

\bibitem{chen2022emphi}
M.~Y. Chen, S.~Li, and Y.~Yang, ``{E}mp{H}i: Generating empathetic responses with human-like intents,'' in \emph{Proceedings of the 2022 Conference of the NAACL: Human Language Technologies}.\hskip 1em plus 0.5em minus 0.4em\relax Seattle, United States: ACL, Jul. 2022, pp. 1063--1074.

\bibitem{gao2021improving}
J.~Gao, Y.~Liu, H.~Deng, W.~Wang, Y.~Cao, J.~Du, and R.~Xu, ``Improving empathetic response generation by recognizing emotion cause in conversations,'' in \emph{Findings of the EMNLP}, 2021, pp. 807--819.

\bibitem{firdaus2021seprg}
M.~Firdaus, U.~Jain, A.~Ekbal, and P.~Bhattacharyya, ``Seprg: sentiment aware emotion controlled personalized response generation,'' in \emph{Proceedings of the 14th International Conference on Natural Language Generation}, 2021, pp. 353--363.

\bibitem{li2016deep}
J.~Li, W.~Monroe, A.~Ritter, M.~Galley, J.~Gao, and D.~Jurafsky, ``Deep reinforcement learning for dialogue generation,'' \emph{arXiv preprint arXiv:1606.01541}, 2016.

\bibitem{saha2020reinforcement}
T.~Saha, S.~Chopra, S.~Saha, and P.~Bhattacharyya, ``Reinforcement learning based personalized neural dialogue generation,'' in \emph{Neural Information Processing: 27th International Conference, ICONIP 2020, Bangkok, Thailand, November 18--22, 2020, Proceedings, Part IV 27}.\hskip 1em plus 0.5em minus 0.4em\relax Springer, 2020, pp. 709--716.

\bibitem{wang2022care}
J.~Wang, Y.~Cheng, and W.~Li, ``Care: Causality reasoning for empathetic responses by conditional graph generation,'' \emph{arXiv preprint arXiv:2211.00255}, 2022.

\bibitem{gu2022hetermpc}
J.-C. Gu, C.-H. Tan, C.~Tao, Z.-H. Ling, H.~Hu, X.~Geng, and D.~Jiang, ``Hetermpc: A heterogeneous graph neural network for response generation in multi-party conversations,'' \emph{arXiv preprint arXiv:2203.08500}, 2022.

\bibitem{zhao-etal-2019-rethinking}
T.~Zhao, K.~Xie, and M.~Eskenazi, ``Rethinking action spaces for reinforcement learning in end-to-end dialog agents with latent variable models,'' in \emph{Proceedings of the 2019 Conference of the NAACL}.\hskip 1em plus 0.5em minus 0.4em\relax Minnesota: ACL, Jun. 2019, pp. 1208--1218.

\bibitem{amutio2015mindfulness}
A.~Amutio, C.~Franco, M.~de~Carmen Pérez-Fuentes, J.~J. Gázquez, and I.~Mercader, ``Mindfulness training for reducing anger, anxiety, and depression in fibromyalgia patients,'' \emph{Frontiers in Psychology}, vol.~5, p. 1572, 2015.

\bibitem{karkar2021understanding}
R.~Karkar, N.~Almuhanna, A.~Althumali, A.~Alkhalifah, R.~Aljamaan, M.~Alsuwaidan, and T.~Alshammari, ``Understanding people’s use of and perspectives on mood-tracking apps: Mixed methods study,'' \emph{JMIR mental health}, vol.~8, no.~8, p. e29368, 2021.

\bibitem{10.5555/3016387.3016435}
I.~V. Serban, A.~Sordoni, Y.~Bengio, A.~Courville, and J.~Pineau, ``Building end-to-end dialogue systems using generative hierarchical neural network models,'' in \emph{Proceedings of the Thirtieth AAAI Conference on Artificial Intelligence}, ser. AAAI'16.\hskip 1em plus 0.5em minus 0.4em\relax AAAI Press, 2016, p. 3776–3783.

\bibitem{park-etal-2018-hierarchical}
\BIBentryALTinterwordspacing
Y.~Park, J.~Cho, and G.~Kim, ``A hierarchical latent structure for variational conversation modeling,'' in \emph{Proceedings of the 2018 Conference of the NAACL}, Louisiana, Jun. 2018, pp. 1792--1801. [Online]. Available: \url{https://aclanthology.org/N18-1162}
\BIBentrySTDinterwordspacing

\bibitem{https://doi.org/10.48550/arxiv.1907.05599}
\BIBentryALTinterwordspacing
T.~Zhao and T.~Kawahara, ``Effective incorporation of speaker information in utterance encoding in dialog,'' 2019. [Online]. Available: \url{https://arxiv.org/abs/1907.05599}
\BIBentrySTDinterwordspacing

\bibitem{radford2019language}
A.~Radford, J.~Wu, R.~Child, D.~Luan, D.~Amodei, and I.~Sutskever, ``Language models are unsupervised multitask learners,'' \emph{OpenAI blog}, 2019.

\bibitem{wu-etal-2020-diverse}
S.~Wu, Y.~Li, D.~Zhang, Y.~Zhou, and Z.~Wu, ``Diverse and informative dialogue generation with context-specific commonsense knowledge awareness,'' in \emph{Proceedings of the 58th Annual Meeting of the ACL}.\hskip 1em plus 0.5em minus 0.4em\relax Online: ACL, Jul. 2020, pp. 5811--5820.

\bibitem{zhang-etal-2020-dialogpt}
Y.~Zhang, S.~Sun, M.~Galley, Y.-C. Chen, C.~Brockett, X.~Gao, J.~Gao, J.~Liu, and B.~Dolan, ``{DIALOGPT} : Large-scale generative pre-training for conversational response generation,'' in \emph{Proceedings of the 58th Annual Meeting of the ACL: System Demonstrations}.\hskip 1em plus 0.5em minus 0.4em\relax Online: ACL, Jul. 2020, pp. 270--278.

\bibitem{zheng-etal-2021-comae}
\BIBentryALTinterwordspacing
C.~Zheng, Y.~Liu, W.~Chen, Y.~Leng, and M.~Huang, ``{C}o{MAE}: A multi-factor hierarchical framework for empathetic response generation,'' in \emph{Findings of the Association for Computational Linguistics: ACL-IJCNLP 2021}.\hskip 1em plus 0.5em minus 0.4em\relax Online: Association for Computational Linguistics, Aug. 2021, pp. 813--824. [Online]. Available: \url{https://aclanthology.org/2021.findings-acl.72}
\BIBentrySTDinterwordspacing

\bibitem{qi-etal-2021-prophetnet}
W.~Qi, Y.~Gong, Y.~Yan, C.~Xu, B.~Yao, B.~Zhou, B.~Cheng, D.~Jiang, J.~Chen, R.~Zhang, H.~Li, and N.~Duan, ``{P}rophet{N}et-{X}: Large-scale pre-training models for {E}nglish, {C}hinese, multi-lingual, dialog, and code generation,'' in \emph{Proceedings of the 59th Annual Meeting of the ACL and the 11th IJCNLP: System Demonstrations}.\hskip 1em plus 0.5em minus 0.4em\relax Online: ACL, Aug. 2021, pp. 232--239.

\bibitem{chen-etal-2022-dialogved}
W.~Chen, Y.~Gong, S.~Wang, B.~Yao, W.~Qi, Z.~Wei, X.~Hu, B.~Zhou, Y.~Mao, W.~Chen, B.~Cheng, and N.~Duan, ``{D}ialog{VED}: A pre-trained latent variable encoder-decoder model for dialog response generation,'' in \emph{Proceedings of the 60th Annual Meeting of the ACL}.\hskip 1em plus 0.5em minus 0.4em\relax Dublin, Ireland: ACL, May 2022, pp. 4852--4864.

\bibitem{Sabour_Zheng_Huang_2022}
S.~Sabour, C.~Zheng, and M.~Huang, ``Cem: Commonsense-aware empathetic response generation,'' \emph{Proceedings of the AAAI Conference on Artificial Intelligence}, vol.~36, no.~10, pp. 11\,229--11\,237, Jun. 2022.

\bibitem{VANDERLEE2021101151}
C.~{van der Lee}, A.~Gatt, E.~{van Miltenburg}, and E.~Krahmer, ``Human evaluation of automatically generated text: Current trends and best practice guidelines,'' \emph{Computer Speech \& Language}, vol.~67, p. 101151, 2021.

\end{thebibliography}

\newpage
\begin{wrapfigure}{l}{2cm}
\centering
\includegraphics[width=2cm,height=2cm]{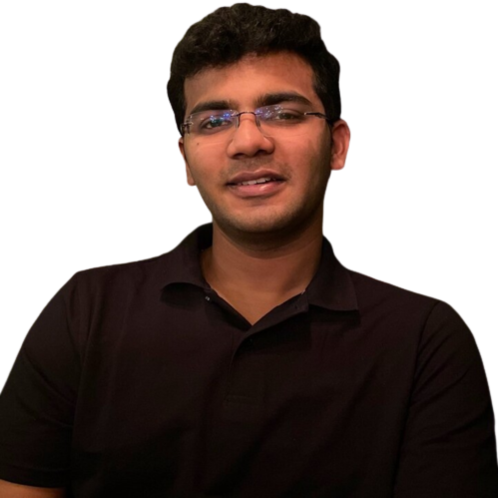}
\end{wrapfigure} 

\noindent \textbf{Aseem Srivastava} is a PhD student at IIIT-Delhi, India. Aseem's primary research interests include Natural Language Processing for Dialogue Systems especially focusing on Virtual Mental Health Assistants (VMHAs) for client's behavioral understanding and response generation.\\

\begin{wrapfigure}{l}{2cm}
\centering
\includegraphics[width=2cm,height=2cm]{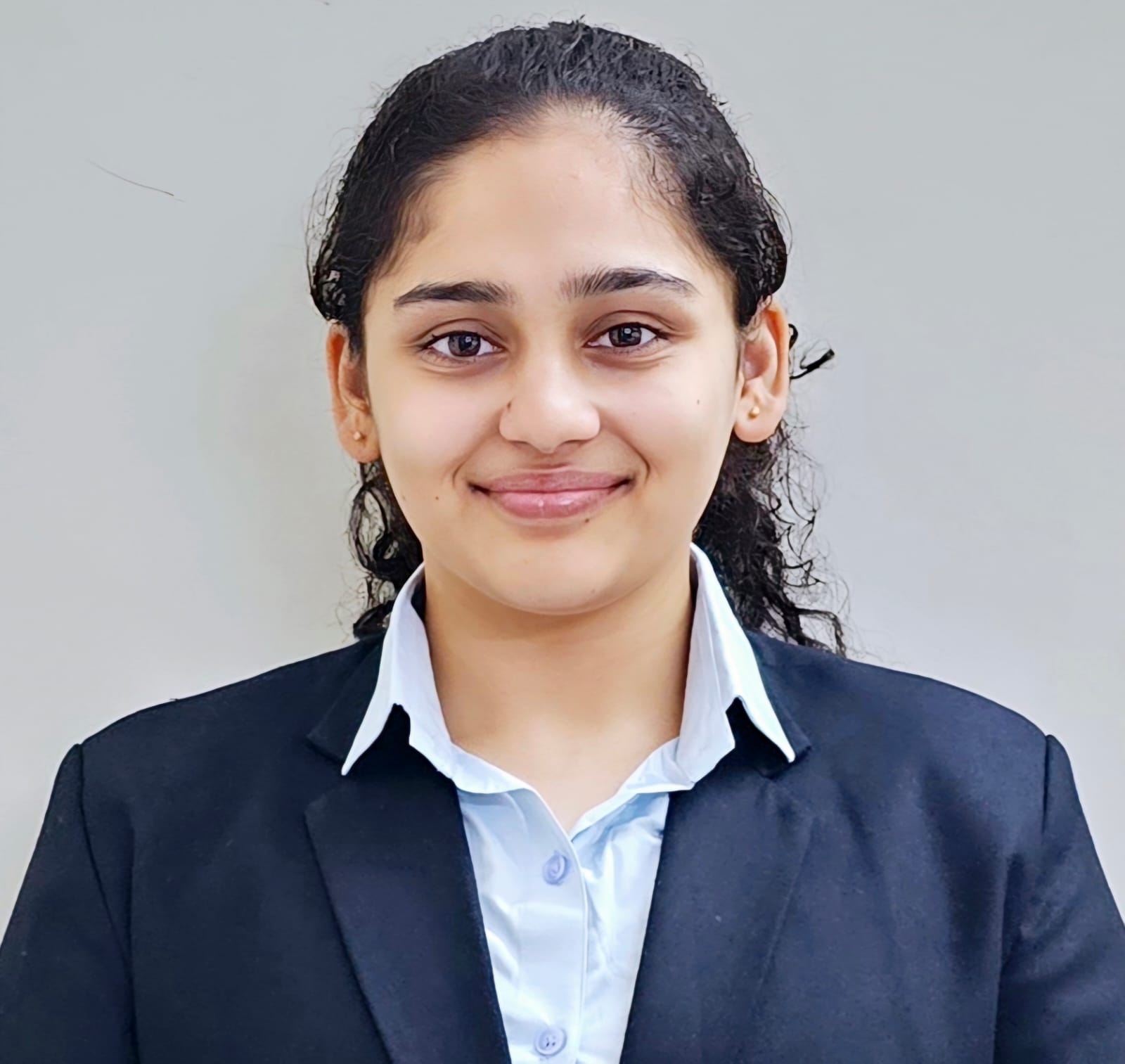}
\end{wrapfigure} 

\noindent \textbf{Gauri Naik} is a Research Associate at IIIT-Delhi. Prior to this, she worked as a Software Developer at Wipro Technologies. Her primary research interests include Natural Language Processing, Computer Vision and Machine Learning. \\

\begin{wrapfigure}{l}{2cm}
\centering
\includegraphics[width=2cm,height=2cm]{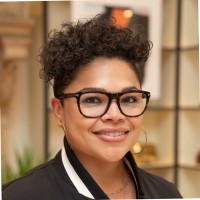}
\end{wrapfigure} 

\noindent \textbf{Alison Cerezo} is a licensed psychologist and a faculty member in the UCSB Counseling, Clinical and School Psychology doctoral program. Her primary line of research centers on addressing social and health disparities using an intersectionality framework.\\

\begin{wrapfigure}{l}{2cm}
\centering
\includegraphics[width=2cm,height=2cm]{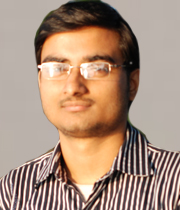}
\end{wrapfigure} 

\noindent   {{\bf Tanmoy Chakraborty}}{\space} is an Associate Professor in the Dept. of Electrical Engineering, Indian Institute of Technology Delhi, India. His broad research interests include Natural Language Processing, Graph Neural Networks, and Social Computing. He is a senior IEEE member and an ACM Distinguished Speaker.
More details about him can be found at \url{tanmoychak.com}.

\begin{wrapfigure}{l}{2cm}
\centering
\includegraphics[width=2cm,height=2cm]{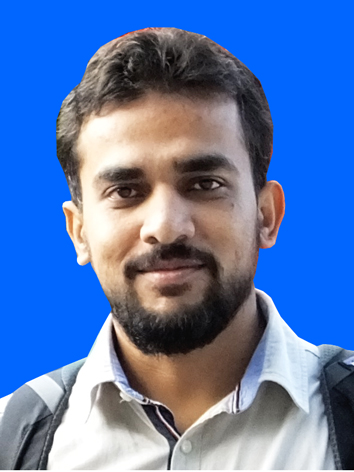}
\end{wrapfigure} 

\noindent \textbf{Md Shad Akhtar} currently holds the position of Assistant Professor at the IIIT-Delhi. His research primarily revolves around Natural Language Processing (NLP) and Deep Learning (DL), with a specialized focus on affective dialogue systems. His received his doctoral degree at the IIT Patna. \\

\end{document}